\documentclass[5pt,twocolumn,letterpaper]{article}

\usepackage{amssymb}
\usepackage{amsmath}
\usepackage{subcaption}
\usepackage{graphicx}

\usepackage{pifont}
\newcommand*{\FeatureTrue}{\ding{52}}
\newcommand*{\FeatureFalse}{\ding{56}}

\usepackage{lineno}

\usepackage[pagebackref=true,breaklinks=true,letterpaper=true,colorlinks,bookmarks=false]{hyperref}

\begin{document}




\title{Synthetic Data for Face Recognition: Current State and Future Prospects}

\author{Fadi Boutros$^{1}$, Vitomir Struc$^{2}$,Julian Fierrez$^{3}$, Naser Damer$^{1,4}$ \\
$^{1}$Fraunhofer Institute for Computer Graphics Research IGD, Darmstadt, Germany\\
$^{2}$ Faculty of Electrical Engineering, University of Ljubljana, Ljubljana, Slovenia \\
$^{3}$ School of Engineering, Universidad Autonoma de Madrid, Madrid, Spain \\
$^{4}$Department of Computer Science, TU Darmstadt,
Darmstadt, Germany\\
Email: fadi.boutros@igd.fraunhofer.de
}
\date{}
\maketitle


\begin{abstract}
\vspace{-3mm}
Over the past years, deep learning capabilities and the availability of large-scale training datasets advanced rapidly, leading to breakthroughs in face recognition accuracy. However, these technologies are foreseen to face a major challenge in the next years due to the legal and ethical concerns about using authentic biometric data in AI model training and evaluation along with increasingly utilizing data-hungry state-of-the-art deep learning models.
With the recent advances in deep generative models and their success in generating realistic and high-resolution synthetic image data, privacy-friendly synthetic data has been recently proposed as an alternative to privacy-sensitive authentic data to overcome the challenges of using authentic data in face recognition development. 
This work aims at providing a clear and structured picture of the use-cases taxonomy of synthetic face data in face recognition along with the recent emerging advances of face recognition models developed on the bases of synthetic data. We also discuss the challenges facing the use of synthetic data in face recognition development and several future prospects of synthetic data in the domain of face recognition. 
\end{abstract}








\vspace{-3mm}
\section{Introduction}

The breakthroughs of deep neural networks and their training optimizations as well as the availability of large-scale identity-labeled face datasets have reshaped the research landscape of face recognition (FR) over the past years. These emerging technologies have dramatically improved FR performances leading to the wider integration of FR in a variety of applications from logical access control and consumer low-end devices to automated border control.
State-of-the-Art (SOTA) FR models \cite{elasticface,ArcFace} utilized large-scale face datasets e.g. CASIA-WebFace \cite{CASIA}, MS-Celeb-1M \cite{MS1M}, or VGGFace2 \cite{VGGFace2} to train deep neural networks (DNN) with millions of trainable parameters, where the goal is to optimize the empirical risk minimization function given input training samples, their corresponding labels, and DNN trainable parameters.   
Achieving such a goal without being over-optimized, i.e. overfitted, requires that training datasets are of large scale (massive number of images of many identities) and representative of various variations that exist in the real world. 
Large and representative data is also required to evaluate FR accuracies against different variations that present in real operation scenarios e.g. pose, aging, occlusion, or lighting. Data is required to evaluate the vulnerability of FR against different types of attacks such as morphing, presentation, master-face, and deep fake attacks.
FR components, face processing models, attack detectors, and face image quality estimation models are not different as they require face data for training and evaluation.
Besides the technical limitation of collecting large-scale data with realistic variations, there are increased concerns about collecting, maintaining, redistributing, and using biometric data due to legal, ethical, and privacy concerns \cite{eu-2016/679}. Consequently, many widely used datasets for FR development such as VGGFace2 \cite{VGGFace2} and MS-Celeb-1M \cite{MS1M} have been retracted by their creator. Table \ref{tab:datasets} summarizes the most widely used datasets to train FR models. Even though many of these datasets have been publically released, there are not any more accessible.

Processing biometric data is governed by a set of legal restrictions \cite{eu-2016/679}. 
Taking the General Data Protection Regulation (GDPR) \cite{eu-2016/679} as an example, it categories biometric data as a special category of personal data subjected to rigorous data protection rules \cite{art9gdpr}, requiring high protection in connection with fundamental rights and freedoms of individuals.
Dealing with such data requires adherence to one of the exemptions of biometric data processing \cite{art9_2gdpr}, the related national laws \cite{art9_4gdpr}, maintaining processing records \cite{art30gdpr}, and the preparation of data protection impact assessment \cite{art35_3gdpr,art37_1gdpr}, among other restrictions. 
Depending on the purpose of the biometric data processing, this set of restrictions can be rigorously extended \cite{art22_4gdpr,art27_2gdpr,art6_4gdpr}.
Besides the legal complications of using and sharing biometric data, ethical requirements are commonly necessary, such as the approval of an ethics committee or competent authorities.

The increased concerns about the legal and ethical use of authentic data in biometrics along with the technical limitation in collecting large and diverse face datasets motivate recent works to propose the use of synthetic data as an alternative to privacy-sensitive authentic data in FR training \cite{SFace,SynFace,USynFace}.
In an attempt to provide a clear understanding of the feasibility of utilizing synthetic face data to train, evaluate, attack, or privacy enhancement, this work is the first to analyze the properties needed of the synthetic data for FR, the use-cases taxonomy of synthetic data in FR, the current state of synthetic-based FR, the limitations and challenges facing the use of current synthetic face data in FR, and possible future research directions that might give a larger space for synthetic data in different aspects of FR development.

\begin{table*}[ht]
\centering
\caption{Overview of the most widely used authentic and synthetic facial datasets commonly used to train FR models, along with the number of images, identities, images per identity, and the fact that each database is public and/or still accessible. Note that many of the public databases are not accessible (raising a practical problem for researchers and developers) anymore based on legal and ethical concerns and even those that are available are ethically questioned as the individual consent of the data subjects is not always insured.}
\resizebox{\textwidth}{!}{%
\begin{tabular}{|c|c|r|r|r|c|c|c|}
\hline
\textbf{Name} & \textbf{Year} & \textbf{\# Images (m)} & \textbf{\# Identities (k)} & \textbf{Avg.} & \textbf{Public} & \textbf{Accessible} & \textbf{Authentic} \\ \hline
CASIA-WebFace \cite{CASIA} & 2014          & 0.5                                                                                 & 10.6                                                                                    & 47                                                                                    & \FeatureTrue    & \FeatureFalse   & \FeatureTrue      \\ \hline
DeepFace \cite{DeepFace}      & 2014          & 4.4                                                                                 & 4.0                                                                                     & 1092                                                                                  & \FeatureFalse   & \FeatureFalse     & \FeatureTrue    \\ \hline
FaceNet \cite{FaceNet}      & 2015          & 200.0                                                                               & 8,000.0                                                                                 & 25                                                                                    & \FeatureFalse   & \FeatureFalse   & \FeatureTrue      \\ \hline
Facebook \cite{FBDataset2}     & 2015          & 500.0                                                                               & 10,000.0                                                                                & 50                                                                                    & \FeatureFalse   & \FeatureFalse   & \FeatureTrue      \\ \hline
VGGFace \cite{VGGFace}      & 2015          & 2.6                                                                                 & 2.6                                                                                     & 992                                                                                   & \FeatureTrue    & \FeatureFalse    & \FeatureTrue     \\ \hline
CelebFaces \cite{CelebFaces}    & 2016          & 0.09                                                                                & 5.4                                                                                     & 16                                                                                    & \FeatureTrue    & \FeatureTrue     & \FeatureTrue     \\ \hline
MS-Celeb-1M \cite{MS1M}   & 2016          & 10                                                                                  & 100.0                                                                                   & 100                                                                                   & \FeatureTrue    & \FeatureFalse   & \FeatureTrue      \\ \hline
MegaFace2 \cite{MegaFace2}     & 2017          & 4.7                                                                                 & 672.0                                                                                   & 7                                                                                     & \FeatureTrue    & \FeatureTrue   & \FeatureTrue       \\ \hline
UMDFaces \cite{UMDFaces}     & 2017          & 0.4                                                                                 & 8.3                                                                                     & 46                                                                                    & \FeatureTrue    & \FeatureFalse   & \FeatureTrue      \\ \hline
VGGFace2 \cite{VGGFace2}     & 2018          & 3.3                                                                                 & 9.1                                                                                     & 363                                                                                   & \FeatureTrue    & \FeatureFalse   & \FeatureTrue      \\ \hline
IMDbFace \cite{IMDBFace}     & 2018          & 1.7                                                                                 & 59.0                                                                                    & 29                                                                                    & \FeatureTrue    & \FeatureTrue    & \FeatureTrue      \\ \hline
MS1MV2 \cite{ArcFace,MS1M}       & 2019          & 5.8                                                                                 & 85.0                                                                                    & 68                                                                                    & \FeatureTrue    & \FeatureFalse     & \FeatureTrue    \\ \hline
MillionCelebs \cite{MillionCelebs} & 2020          & 18.8                                                                                & 636.0                                                                                   & 30                                                                                    & \FeatureFalse   & \FeatureFalse    & \FeatureTrue     \\ \hline
WebFace260M \cite{WebFace260M}  & 2021          & 260                                                                                 & 4,000.0                                                                                 & 65                                                                                    & \FeatureTrue    & \FeatureTrue     & \FeatureTrue     \\ \hline
WebFace42M \cite{WebFace260M}   & 2021          & 42                                                                                  & 2,000.0                                                                                 & 21                                                                                    & \FeatureTrue    & \FeatureTrue    & \FeatureTrue      \\ \hline \hline

SynFace \cite{SynFace}   & 2021          & 0.5                                                                                  & 10                                                                                 & 50                                                                                    & \FeatureTrue    & \FeatureTrue    & \FeatureFalse     \\ \hline
DigiFace-1M-A \cite{digiface1m}   & 2022          & 0.72                                                                                  & 10                                                                                 & 72                                                                                    & \FeatureTrue    & \FeatureTrue    & \FeatureFalse     \\ \hline
DigiFace-1M-B \cite{digiface1m}   & 2022          & 0.5                                                                                  & 100                                                                                 & 5                                                                                    & \FeatureTrue    & \FeatureTrue    & \FeatureFalse     \\ \hline

SFace \cite{SFace}   & 2022          & 0.63                                                                                  & 10.6                                                                                 & 60                                                                                    & \FeatureTrue    & \FeatureTrue   & \FeatureFalse      \\ \hline
USynthFace \cite{USynFace}   & 2022          & 0.4                                                                                  & 0.4                                                                                 & 1                                                                                    & \FeatureTrue    & \FeatureTrue   & \FeatureFalse      \\ \hline
\end{tabular}
}
\label{tab:datasets}
\end{table*}

\section{Where is the synthetic data used?}
\label{sec:usecase}

To analyse the properties of the needed synthetic data, one should start by building a clear taxonomy of the different possible uses-cases of synthetic data in its interaction with FR.
This taxonomy here will consider the operations where the synthetic data is used to interact with the recognition part of FR systems, i.e. the feature extraction.
Therefore, synthetic data that is meant to interact with other system components, as defined in ISO ISO/IEC 19795-1:2021 \cite{iso_metric}, are out of scope, e.g. synthetic data used to train or evaluate face detection or segmentation solutions. 
Additionally, synthesizing faces as a means of domain transformation, e.g. from thermal to visible face appearance \cite{DBLP:conf/icb/MallatDBKD19} is also out of scope as it just transfers the appearance of the image.

\begin{figure*}[ht!]
\begin{center}
\includegraphics[width=0.88\linewidth]{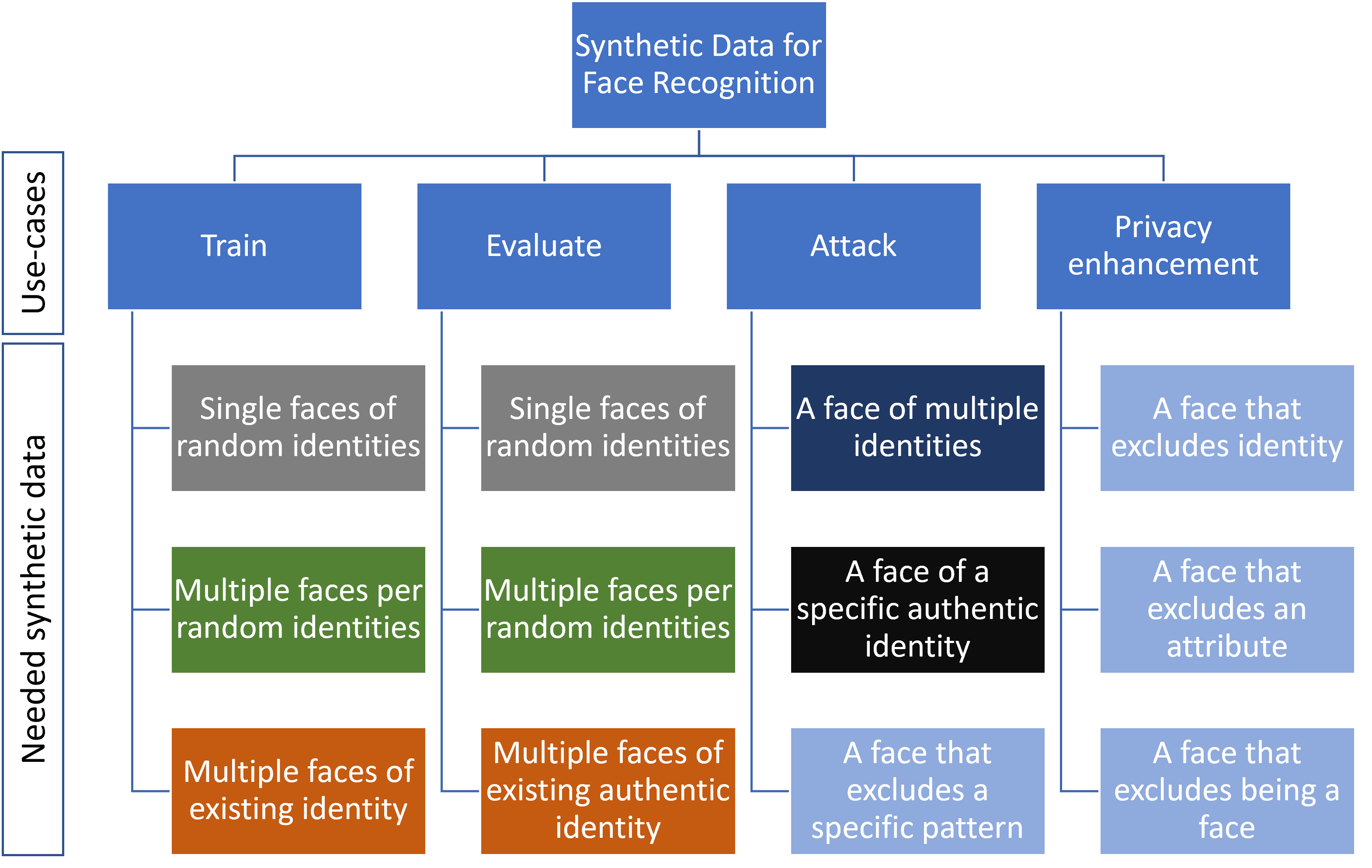}
\end{center}
\vspace{-0mm}
\caption{A taxonomy of the synthetic data use-cases (on the top of the figure) directly interacting with FR models, either by training them, evaluating them, attacking them, or enhancing the privacy of the information extracted by them. This taxonomy lists the existing and foreseen synthetic data types that are needed by these use-cases (under each use-case). These data needs are grouped by their main properties by color and discussed, along with the use-cases in this paper. }
\label{fig:tax}
\vspace{-0mm}
\end{figure*}

Figure \ref{fig:tax} presents the use-case taxonomy of the synthetic face data interaction with FR. 
These use-cases are categorised under 4 groups, along with the properties of the possibly needed data under each category (the latter will be discussed in detail in the next section).
The four use-case categories are discussed in the following.

\begin{enumerate}
    \item \textbf{Training FR:}
    Modern FR solutions are based on deep learning models that are either trained directly to generate identity-discriminant feature representations (e.g. triplet loss \cite{FaceNet}) or to classify the identity classes in the training data (e.g. ArcFace \cite{ArcFace}, ElasticFace \cite{elasticface}, etc.). 
    In the latter approach, embeddings proceeding the classification layer of the network are then used to extract the identity-discriminant representations. This family of approaches is currently predominantly leading to SOTA FR performances.
    In both cases, training face data that represents the high inter and intra-class diversity of real applications is needed to train the models.
    As mentioned in the introduction, the diversity of such data, if authentic, is limited by practical data collection constraints, and  its collection and handling are hedged by privacy, legal, and ethical concerns. 
    Synthetic data can come in handy to train such FR models in different manners based on the training requirements.
    If the model is trained in one of the two approaches mentioned above, then the synthetic data has to contain a large number of identities and multiple samples of each identity.
    If the model is trained on partially authentic data, however, the intra-class variation of this data is low, then the synthetic data needs to contain multiple samples for each of the authentic identities, i.e. act as an augmentation strategy.
    Finally, if the FR model is trained in an unsupervised manner, then the synthetic training data is not largely concerned with the identity grouping, but rather just requires a set of faces of random identities.
    This data has also been shown to be successful in training processes during the training-aware quantization of models based on full precision parameters \cite{DBLP:conf/icpr/BoutrosDK22}.
    Although it is out of the scope of this work, synthetic faces of this kind can also be used to train face detectors, face segmentation, and attack detection methods (e.g. morphing attack detection \cite{DBLP:conf/cvpr/DamerLFSPB22}).
    \item \textbf{Evaluating FR:}
    FR algorithmic evaluation, following the ISO ISO/IEC 19795-1:2021 \cite{iso_metric}, requires the existence of a large set of genuine (same identity) and imposter (different identity) face image pairs that represent the real operational scenario.
    The need for a large number of these pairs is intensified by the ever-more accurate performance of FR algorithms.
    FR algorithms can produce two main algorithmic errors, genuine pairs classified wrongly as imposters (false non-match (FNM)) or imposter pairs classified wrongly as genuine (false match FM).
    As the algorithms produce lower and lower rates of decision errors, the FM rates (FMR) and FNM rates (FNMR), the number of evaluated pairs required to produce statistically significant evaluation results become higher.
    This need for large-scale evaluation data is one of the main motivations behind requiring synthetic data for the evaluation.
    Another reason is that some authorities that require in-house testing on their own data when purchasing FR solutions do only possess a single image per identity in their databases (think of visa systems) and thus it is impossible to have genuine pairs to evaluate FR algorithms.
    Such situations would require synthetic data to be generated so it belongs to a certain authentic identity, but with realistic variations.
    In a third scenario where the operation scenario would require a very low FMR, the need for a huge number of imposter pairs is required to evaluate, with statistical significance, the FMR. 
    In such cases, random synthetic faces with random identities can be used to create such imposter pairs.
    Again, although it is out of the scope of this work, these synthetic faces, regardless of their identity information, can be used to evaluate face detectors, face segmentation, and presentation/morphing attack detection.
    \item \textbf{Attacking FR:} 
    Commonly, developers would use technology to enhance the convenience and security of individuals and societies. 
    However, technology can also be used maliciously to create attacks on individuals, systems, and societies.
    This is the case also with synthetic face data, which can also be used as an attack.
    Synthetic data can be created so that a certain face can be matched with two or more faces.
    This can target automatic FR comparison or human image verification, or both.
    Such attacks can be face morphing attacks, where an image is generated to match two or more identities, then used on an identity or travel document with the alphanumeric data of when the targeted matches. 
    Later such a document can be used by the other targeted identities illegally, leading to a serious security threat.
    Another attack in the same category is the MasterFace attack, where the synthetic face is created to match a wider proportion of the population, raising many security threats.
    The second type of attack by generated face images might focus on generating a face image of a specific identity.
    Such attacks are commonly referred to as Deep-Fakes and they are commonly used to fool the viewer into wrongly believing that a certain person has said or done an action in an image or a video.
    A third attack can use synthetic faces that maintain a certain identity but excludes a specific pattern with the aim of attacking a biometric-based system that ensures a legal operation of a process.
    Such an attack can be by presenting the attacker's real identity, but excluding the information that points out that the user is underage, in a service that requires age verification.    
    \item \textbf{Enhancing the privacy for FR users:}
    Although excluding certain patterns from generated images of specific identities can be seen as an attack on biometric systems, in different use-cases, they can be seen as a privacy-enhancing tool when they are used to avoid the illegal or unconsented processing of the data.
    Such generation of the data aims at maintaining a certain set of visual patterns but removing the clues of a specific pattern.
    Depending on the use-case, this excluded pattern can be related to the identity in what is widely known as image-level face de-identification, which is defined under the standard ISO/IEC 20889:2018 \cite{iso_de_identification}.
    The excluded pattern can be related to certain soft biometric attributes like age or gender, which is commonly referred to as soft-biometric privacy enhancement.
    Although it is out of the scope of this work, the generated faces can exclude patterns that makes them detectable to face detection tool, i.e. excluding the information that makes the face a face in the view of automatic face detection.
\end{enumerate}

So far, we presented a discussion on the possible use-cases of synthetic face data in FR. Each of these use-cases has different needs when it comes to synthetic data. These needs are discussed in the next section.

\section{What data is needed and what properties make it good?}%

The properties of the needed synthetic data under the different use-cases (discussed in the previous section) are grouped by their required properties under different colors in Figure \ref{fig:tax} and are discussed in detail in the following:

\begin{enumerate}
    \item \textbf{Single faces of random identities:}
    As detailed in the previous section, and illustrated in Figure \ref{fig:tax}, synthetic face images of random identities without the requirement of multiple images to belong to one identity can be used for training FR models in an unsupervised manner. 
    They additionally can be used to evaluate FR models, specifically evaluate the FMR, especially when the targeted operational point is at a very low FMR, requiring an extremely large number of diverse imposter pairs to make the evaluation result statistically significant.
    Here, such data should be realistic, i.e. act like authentic data when processed by the FR model.
    A successful way to measure that was proposed in \cite{DBLP:conf/icpr/BoutrosDK22} and it is based on comparing the activation function value ranges in the FR model when processing authentic data versus when processing the synthetic data.
    Additionally, the distribution of the comparison scores between pairs of these single images of random identities should theoretically be similar to those of imposter comparisons of authentic data to ensure the similarity to the authentic inter-identity variation, which was explored in \cite{USynFace}.
    \item \textbf{Multiple faces per random identity:}
    This kind of data represents what one would typically expect from FR training or evaluation data.
    That is, multiple identities, with multiple images per identity.
    This, given a sufficient inter and intra-class (identity variation), can be used to train an FR model in a supervised manner.
    This also would contain both imposter and genuine pairs to evaluate the performance of FR by calculating both possible errors, FMR and FNMR.
    Such data should also interact with the FR model similarly to authentic data, this as mentioned earlier can be measured by monitoring the value range of the model's activation functions.
    Here, the data should possess an inter and intra-class variability of the targeted authentic data scenario.
    We specify ``targeted'' here as different evaluation and training goals of FR might occur, e.g. a model is evaluated specifically for cases with an extreme pose or extreme age differences between the comparison pairs (intra-class variations), or for cases of pairs of twins or siblings (inter-class variation). This goes for training as well, as an FR model can be trained to specifically be tolerant to mask occlusions, and thus the training data inter and intra-class diversity should represent that.
    The suitability of such data can be measured by comparing its genuine and imposter comparison scores distributions with that of the targeted authentic data (which can be much smaller in size) as performed in \cite{SFace}.
    For specifically targeted attribute variations, such as age and pose, attribute predictors can be used to ensure the existence of such attribute variations in the synthetic data to the same degree as the authentic data.
    \item \textbf{Multiple faces of an existing identity:} 
    Authentic face data with insufficient intra-class variation is problematic for the training and evaluation of FR.
    In terms of training an FR model, such data will lead to models that are not trained to tolerate intra-class variation (e.g. pose, expressions, age, illumination, etc.) and thus are expected to lead to high FNMR in practical operations.
    When evaluating FR, evaluation data in some practical cases such as an authority that possesses only a single (or few) images per identity (e.g. visa applicant database) would not be sufficient to evaluate the expected FNMR as no (or few) genuine pairs exist in the data. 
    Both cases require acquiring more samples of each of the existing identities. 
    These samples have to be of realistic variation that matches the targeted scenario.
    Such samples might be created synthetically and would act as an augmentation approach when training an FR model, or as additional samples to create genuine pairs when evaluating FR models (or training FR in a triplet loss-like strategy).
    Such synthetic data should interact with the FR model similarly to authentic data, as previously discussed.
    It should also result in genuine comparison score distribution that matches the targeted authentic data scenario.
    One must take notice that this should be the case when the pairs are between the existing authentic sample is compared to the synthetic images of the same identity, but also, if needed, between the synthetically generated samples of the same identity themselves. 
    \item \textbf{A face of multiple identities:}
    A synthetic face can also be used as an attack, the fact that a face can be generated synthetically with properties that enables an attack on identity systems pursues researchers to foresee such attacks.
    A face can be synthesized in a way that it matches two more specific (known) identities to create what is referred to as a morphing attack.
    A morphing attack image is designed to match with a number of specific identities and can be created on the image level by interpolating the images of the targeted identities, or generated synthetically to possess the identity information of the targets \cite{DBLP:conf/btas/DamerS0K18}. 
    Such an image, if used in association with a passport or an identity document can enable multiple persons to be verified to the alphanumeric information on the card.
    A wider attack that surfaced lately in the literature is the MasteFace attack, where the attack image is synthesized to match a wide range of the population without the need to know the targeted identities \cite{DBLP:conf/icb/NguyenYEM20}.
    As these attacks might be used to attack visual inspection, automatic verification, or both, they first have to have a natural appearance.
    This natural appearance is best measured by user studies, where individuals are asked if an image appears realistic or not.
    The vulnerability of automatic FR to such attacks, and thus the measure of how good is the synthetic data for its purpose, can be measured using the Mated Morph Presentation Match Rate (MMPMR) \cite{DBLP:conf/biosig/ScherhagNRGVSSM17}. The MMPMR refers to the fraction of morphs whose similarity to both identities used to morph, are below the selected FR comparison score threshold relative to all morphs.
    \item \textbf{A face of specific authentic identity:}
    Synthesizing a face of a specific authentic identity is usually related to the need to synthesize this face with also a specific expression or domain, unlike generating such faces of an authentic identity where a realistic variation is needed.
    This is commonly related to what is referred to as DeepFake faces but also includes other face manipulation techniques such as expression and attribute manipulations. 
    As such attacks aim at manipulating human viewers, their success is best measured by how realistic they are to these viewers and how well they succeeded in the targeted manipulation in the view of the viewers through user studies related to the exact goal of the manipulation.
    However, more within the scope of this work is the ability of these attacks to fool automatic FR and attack detection algorithms.
    A comprehensive survey on the issue of DeepFakes and facial image manipulation is presented by Tolosana et al. in \cite{deeepfakes}.    
    \item \textbf{A face that excludes a specific pattern:} 
    A face synthesizing process can maintain a subset of patterns from a specific face and excludes other subsets of these patterns.
    Such patterns can be identity information, age, gender, ethnicity, or even the patterns that make a face detectable as a face, among other attribute patterns. 
    Such a process can be seen as an attack if it is aimed at avoiding a consented required process, such as automatic age verification to receive a service or make an online purchase.
    However, such a process can also be seen as a privacy enhancement mechanism. 
    Excluding the identity, while maintaining the image appearance and other attributes to some degree is commonly referred to as image-level face de-identification and it aims at avoiding the unconsented identification of face images, whether in the public or private space.
    A subset of this is to exclude the patterns of the face that makes it detectable and thus avoid further processing.
    Removing other patterns like gender or age falls within the image-level soft-biometric privacy enhancement techniques that aim at maintaining the identification possibilities without allowing unconsented estimation of soft-biometric attributes. 
    Evaluating the ability to synthesize these face images is based on evaluating the degree to which the patterns that need to be excluded and the ones that need to be maintained are detectable, where the first need to be as undetectable as possible and the latter needs to be as detectable as possible.
    A comprehensive survey and discussion on these technologies are presented by Meden et al. in \cite{DBLP:journals/tifs/MedenRTDKSRPS21}.   
\end{enumerate}

\begin{table*}[!ht]
\caption{Verification accuracies (\%) on five different FR benchmarks achieved by the supervised and unsupervised FR models trained on the synthetic training databases with the numbers of real and synthetic training samples.
The result in the first row is reported using the FR model trained on the authentic dataset to give an indication of the performance of an FR model trained on the authentic CASIA-WebFace dataset \cite{CASIA}.
To provide a fair comparison,  all model results are obtained from the original published works using the same network architecture (ResNet50) trained on relatively same training dataset size. KT refers to knowledge transfer from the pretrained FR model. LFW \cite{LFWTech}, AgeDB-30 \cite{AgeDb}, CFP-FP \cite{CFP-FP}, CA-LFW \cite{CALFW}, CP-LFW \cite{CPLFWTech} are widely used FR evaluation benchmarks.}
\label{tab:FR_accuracies}
\centering
\resizebox{\linewidth}{!}{
\begin{tabular}{|c|c|c|c|c|c|c|c|c|c|c|}
\hline
\textbf{Method}               & \textbf{Unsupervised} & \textbf{Data augmentation}  & \textbf{\# Synthetic Images} & \textbf{\# Authentic Images}&  \textbf{KT} & \textbf{LFW}   & \textbf{AgeDB-30} & \textbf{CFP-FP} & \textbf{CA-LFW} & \textbf{CP-LFW} \\ \hline

CosFace \cite{Cosface}        & \FeatureFalse & - & 0 & 500K & \FeatureFalse &99.55 & 94.55   & 95.31 &  93.78 & 89.95 \\ \hline \hline
SynFace \cite{SynFace}       & \FeatureFalse         & GAN-based                                                                                  & 500K  & 0     & \FeatureFalse     & 91.93         & 61.63               & 75.03              & 74.73               & 70.43               \\ \hline

DigiFace-1M   \cite{digiface1m} & \FeatureFalse         & -                                                                                        & 500K     & 0        &  \FeatureFalse                      & 88.07              & 60.92                 & 70.99                & 69.23                & 66.73                           \\  \hline
DigiFace-1M \cite{digiface1m} & \FeatureFalse         & Accessory + Geometric and color                                                                                       & 500K & 0 & \FeatureFalse           & 95.40          & 76.97                & 87.40               & 78.62               & 78.87               \\ \hline
SFace   \cite{SFace}       & \FeatureFalse         & -                                                                                 & 634K   & 0   & \FeatureFalse       & 91.87          & 71.68  & 73.86           & 77.93  & 73.20           \\ \hline

USynthFace   \cite{USynFace}          & \FeatureTrue          & GAN-based + Geometric and color                                                                                    & 400K           & 0  & \FeatureFalse & 92.23 & 71.62             & 78.56 & 77.05           & 72.03           \\ \hline 
IDnet \cite{CVPRW_JAN23} & \FeatureFalse & - & 528K            & 0  &    \FeatureFalse                 & 84.83              & 63.58                 & 70.43                & 71.50                & 67.35                             \\ \hline

IDnet \cite{CVPRW_JAN23}  & \FeatureFalse & Geometric and color  & 528K            & 0  &    \FeatureFalse                    & 92.58              & 73.53                 & 75.40                & 79.90 (3)            & 74.25                              \\
\hline \hline

SynFace  \cite{SynFace}      & \FeatureFalse         & GAN-based                                                                                  & 500K  & 40K    & \FeatureFalse      & 97.23         & 81.32               & 87.68              &85.08              & 80.32             \\ \hline
DigiFace-1M \cite{digiface1m} & \FeatureFalse         & Accessory + Geometric and color                                                                                       & 500K & 40K      & \FeatureFalse      & 99.05          & 89.77               & 94.01              & 90.08              & 87.27               \\ \hline
SFace    \cite{SFace}      & \FeatureFalse         & -                                                                                 & 634K   & 0     & \FeatureTrue     & 99.13          & 91.03  & 91.14            &  92.47  & 87.03           \\ \hline

\end{tabular} }
\end{table*}

\section{Where are we now?} 

\subsection{Face image generation:} A deep generative model (DGM) is a deep neural network that is trained to interpret and model a probability distribution of the authentic training data. 
Specifically, a deep generative model takes random points from e.g. Gaussian distribution and maps them through a neural network such as the generated distribution closely matches the authentic data distribution.
The main DGM approaches that are proposed in the literature are Variational Auto-Encoder (VAE) \cite{vae}, Generative Adversarial Network (GAN) \cite{gan}, Autoregressive model \cite{pixelrnn}, and Normalizing Flows \cite{normalizing_flow} and Diffusion Models (DiffModel) \cite{diffusion}, in addition to a large number of hybrid models that combined two of previous approaches such as GAN with VAE \cite{DBLP:conf/eccv/KowalskiGEBJS20}. A comprehensive review of deep generative modelings is presented by \cite{DBLP:journals/pami/Bond-TaylorLLW22}.
Each of these approaches presented contributions towards providing a better trade-off between generated sample quality i.e. producing samples of high perceived quality and fidelity that resemble the DGM training data, inference time i.e. enabling fast sampling mechanism, architecture restrictions i.e. some of the DGMs are limited to underlying network architecture and sample appearance variations.

\subsection{How do the DGM approaches match the needed synthetic face data properties?}
\begin{itemize}
    \item Single faces of random identities: DGM approaches such as StyleGAN \cite{styleGAN} presented very promising results in generating single faces of random synthetic identities with high visual fidelity. However, the generated faces could share the identity information, to a small degree, with DGM's original training (as reported in \cite{DBLP:conf/wacv/TinsleyCF21,SFace}).
    \item Multiple faces per random identities: Approaches such as Face-ID-GAN \cite{DBLP:conf/cvpr/Shen0YWT18}, DiscoFaceGAN \cite{DBLP:conf/cvpr/DengYCWT20}, GAN-Control \cite{DBLP:conf/iccv/ShoshanBKM21}, InterFaceGAN \cite{DBLP:journals/pami/ShenYTZ22}, and CONFIG \cite{DBLP:conf/eccv/KowalskiGEBJS20} proposed GAN models based on disentangled representation learning to conditionally generate face images from synthetic identities with predefined attributes e.g. age, pose, illumination, or expression.  As generated images are explicitly controlled by a predefined set of attributes, such images might lake the intra-class diversity that exists in real-world face data and it is needed to train and evaluate FR. 
    \item Multiple faces of an existing identity: DGM approaches such as CONFIG \cite{DBLP:conf/eccv/KowalskiGEBJS20} are able to regenerate multiple faces of an existing identity by reconstructing input faces with a predefined set of  attributes such as changing expression, wearing sunglasses, adding makeup, or changing hair color.   However, such attribute manipulation approaches might induce some artifacts in reconstructed faces, which might affect identity preservation between the input and the reconstructed faces. Also, as such approaches are explicitly manipulating the attributes of their input faces, the generated faces might not contain large appearance variations, which are needed to train and evaluate FR models. More importantly, identity preservation in reconstructed samples is rarely evaluated and reported. 
    \item A face of multiple identities: DGM approaches were not explicitly designed and trained to generate a face of multiple identities. However, recent works such as MorGAN \cite{DBLP:conf/btas/DamerS0K18}, MIPGAN \cite{MIPGAN}, and MorDIFF \cite{DBLP:conf/iwbf/Damer23},  make use of generative models to generate a face of multiple identities by interpolating two or more latent vectors of synthetic or real faces and then generating a new face of multiple identities. In a similar manner, however, with latent vector optimization rather than optimization, MasterFaces \cite{DBLP:conf/icb/NguyenYEM20} are generated to match unknown identities. 
    \item A face of specific authentic identity: DGM approaches that targeted image-to-image modeling achieved impressive results in generating a face of specific authentic identity. This has been commonly achieved by manipulating the input source face to match specific attributes or a target domain while maintaining the identity information of the source image. Although such approaches did not target generating Deep-Fake attacks, they have been widely used in generating such kinds of attacks \cite{deeepfakes}.
    \item A face that excludes a specific pattern: None of the SOTA DGM  approaches explicitly target generating a face that excludes a specific pattern. A number of works make use of DGM approaches to exclude a specific pattern e.g. identity, age, or gender of authentic input faces, especially when such models include attribute disentanglement. However, to the best of our knowledge, none of the previous works present solutions to generate a face of synthetic identity that excludes a specific pattern, rather this is done for faces of authentic identities. An overview of the current state of this issue can be found in \cite{DBLP:journals/tifs/MedenRTDKSRPS21}.
\end{itemize}

\begin{figure}[ht!]
\begin{center}
\includegraphics[width=0.9\linewidth]{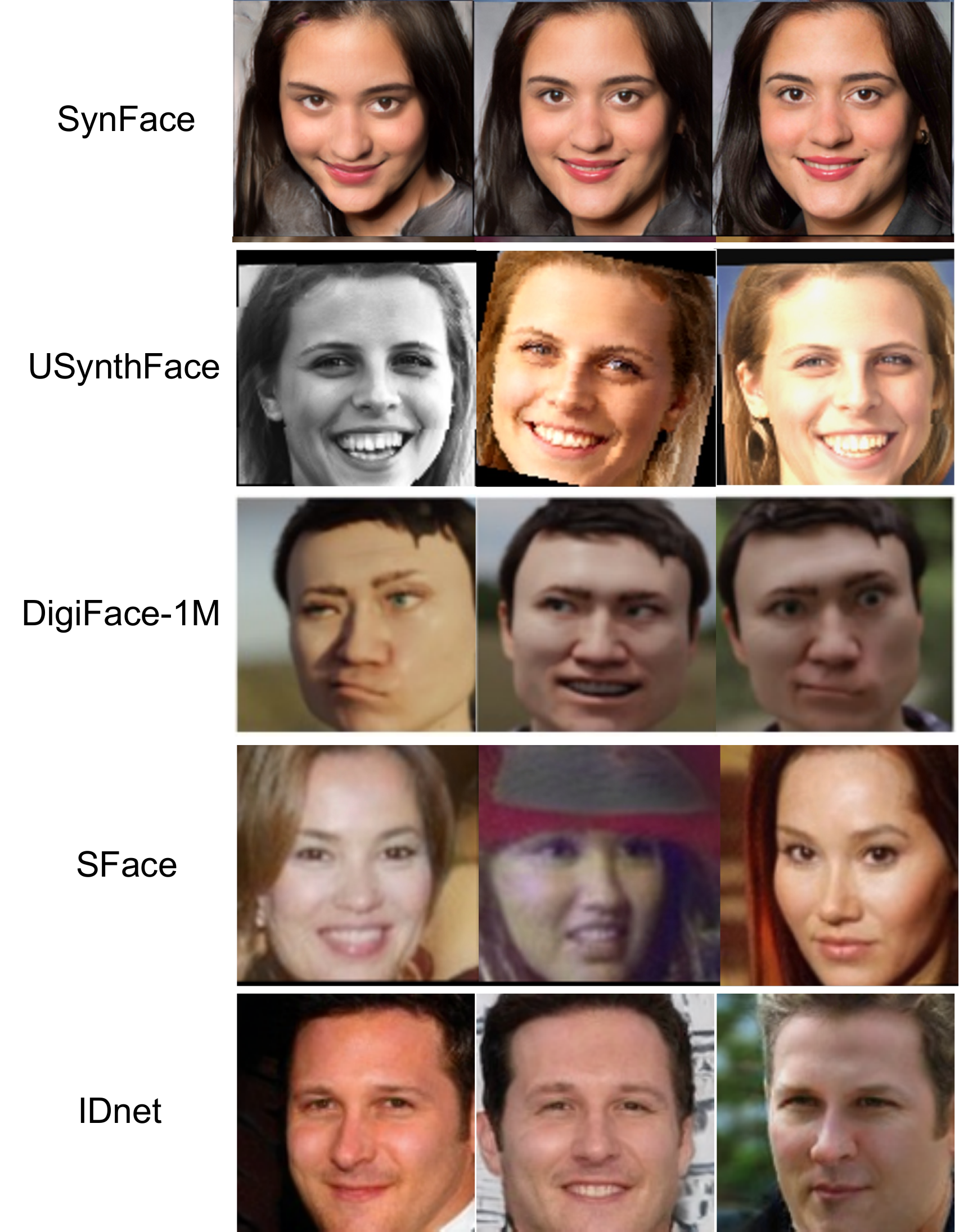}
\end{center}
\vspace{-0mm}
\caption{ Sample of synthetic data used in SynFace \cite{SynFace}, UsynthFace \cite{USynFace}, DigiFace-1M \cite{digiface1m} SFace \cite{SFace} and IDnet \cite{CVPRW_JAN23}. It can be clearly noticed the high variations in SFace images in comparison to other synthetic datasets. Although SynFace and UsynthFace utilized the same DGM (DiscoFaceGAN), it can be also observed the appearance variations in USynthFace using geometric and color transformations. }
\label{fig:samples}
\vspace{-0mm}
\end{figure}

\subsection{What is the current state of the defined use-cases?} 
Very recently a few works build on existing DGM approaches to propose FR based on synthetic data. The following discussion presents the use of synthetic data in FR grouped by the use-cases (discussed earlier in this paper and presented in Figure \ref{fig:tax}).

\subsubsection{Training FR}
Recently, synthetically generated face data has been proposed as an alternative to privacy-sensitive authentic data to train FR models mitigating the technical, ethical, and legal concerns of using authentic biometric data in training FR models.
The currently proposed approaches in the literature utilized synthetically generated data to train unsupervised (UsynthFace \cite{USynFace}) or supervised FR models (SFace\cite{SFace}, SynFace\cite{SynFace}, DigiFace-1M\cite{digiface1m} and IDnet \cite{CVPRW_JAN23}). 
Training the unsupervised FR model as in UsynthFace requires that the training data maintain the property 1 (Section \ref{sec:usecase}) i.e. single face of random identities, while supervised approaches, SFace,  SynFace, IDnet, and DigiFace-1M, require that the training data maintain the property 2 i.e. multiple faces per random identities (Section \ref{sec:usecase}).
Some of these approaches, SynFace and DigiFace-1M,  proposed combining authentic with synthetic data during the training or transferring the knowledge from the pretrained FR model to improve the recognition accuracies. Others (USynthFace) utilized only synthetic data for FR training. Most synthetic FR approaches utilized GAN-based (UsynthFace, SynFace) and/or geometric and color transformation data augmentation (UsynthFace, IDnet, and DigiFace-1M) methods to create more challenging training samples improving the model recognition accuracies.  Table \ref{tab:FR_accuracies} summarizes the achieved accuracies on five FR benchmarks by recent FR models trained on synthetic data. 
It can be observed from the reported results in Table \ref{tab:FR_accuracies} that including data augmentation in FR model training significantly improved the recognition accuracies. Also,  the unsupervised FR model (UsynthFace \cite{USynFace}) obtained very competitive results using unlabeled data to supervised synthetic-based FR models.

\subsubsection{Evaluating FR} A few works proposed the use of synthetic data for evaluating FR. 
SynFace \cite{SynFace} presented a synthetic version of the Labeled Faces in the Wild (LFW) dataset \cite{LFWTech} and evaluated two FR models trained on authentic and synthetic data, respectively on the synthetic version of the LFW. 
The model trained on real data achieved an accuracy of 98.85\% and the one trained on synthetic data achieved an accuracy of 99.98\%. 
The work \cite{SynFace} also suggested that the degradation in the verification performance between the two models is due to the domain gap between synthetic and real training images.

\subsubsection{Attacking FR} DGM approaches have been widely and successfully utilized to generate morphing, MasterFace, deep-fake, and manipulation attacks on FR.
Researchers generally attempt to foresee such attacks and evaluate their potential.
Deep-fake and face manipulation attacks are already a serious problem facing modern societies and their generation is becoming more available and realistic with time \cite{deeepfakes}. Morphing attacks based on synthesized faces are a serious threat and FR recognition vulnerability to them is getting close to that of image-level morphing \cite{MIPGAN}. MasterFace attacks are relatively new, their initial proposed form is based on optimization on a relatively weak FR model \cite{DBLP:conf/icb/NguyenYEM20} with other works arguing their feasibility \cite{DBLP:conf/icb/terh22}.
However, on the other hand, synthetic data has helped create privacy-friendly databases for the detection of such attacks, specifically, the morphing attack \cite{DBLP:conf/cvpr/DamerLFSPB22,SYN_MAD} and  face presentation attack \cite{CVPRW_FANG23}. Huber et al. \cite{SYN_MAD} organized a competition on face morphing attack detection (MAD) based on privacy-aware synthetic training data \cite{DBLP:conf/cvpr/DamerLFSPB22}. The competition aimed at promoting the use of synthetic data to develop MAD solutions and attracted 12 solutions from both academia and industry.

\subsubsection{Privacy enhancement}
Main advances in this respect are presented under one of two categories, de-identification or soft-biometric privacy. De-identification can be achieved by adding adversarial noise to the image, image obfuscation, and image synthesis, the latter being the core focus of this work.
Many solutions have been proposed in the literature, with a recent overview of these solutions presented in \cite{DBLP:journals/tifs/MedenRTDKSRPS21}.
The main challenge so far in this domain is the cross-FR model performance as most works showed very good performances on the FR models that were used to optimize the solution, however, this performance drops when using other unknown FR models.
Syntheses-based soft-biometric privacy followed a similar trend as de-identification, however, with much less dominance in the literature.
In this aspect, many works rather focused on soft-biometric privacy on the template level rather than the image. Image and template level techniques are surveyed in \cite{DBLP:journals/tifs/MedenRTDKSRPS21}. An example of image-based techniques is the FlowSAN \cite{FlowSAN} aimed at minimizing gender information in the resulting images. Here, as the target is the soft-biometrics and not the identity, the main challenge is to achieve generalized performance across soft-biometric estimators while maintaining FR performance across FR models.

\section{Where can we do better?}
Here, based on the discussed use-cases taxonomy, the synthetic data requirements, and their current state along with the generation process, we discuss the main issues where further improvement in future research can have a strong effect on the use of synthetic data in FR. The following discussion will touch on the generation process, the defined use-cases, as well as the general lack of well-defined suitability evaluation protocols.

\subsection{Face image generation} Generating realistic and high-quality samples along with enabling high sampling speed and high-resolution scaling have derived the main contributions of recent generative models proposed in the literature. 
In addition, some DGM approaches targeted specific applications such as image in-painting, attribute manipulation, face aging, image super-resolution, and image-to-image and text-to-image translations. 
Such applications mainly require that the generated samples are of high visual fidelity with less focus on the identity information, which might be less optimal for biometric applications.
When developing DGM for FR use-cases, the solution should focus on the utility of the generated images for the given tasks rather than only focusing on the human-perceived quality.
The emerging works on training FR solutions, presented earlier, are considered the first step in this regard.
This focus on utility, rather than only the perceived quality, should be the main drive in future research when synthesizing images for FR.

\subsection{Training FR}
Recent works that proposed the use of synthetic face data for FR utilized deep neural network architectures with hyper-parameters that are optimized on authentic data. 
Such training paradigms might be sub-optimal for learning face representations from synthetic data. Future research works might target proposing network architectures or training paradigms designed specifically to learn from synthetic data.
In general, training FR solutions of synthetic data still fails behind those trained on authentic data in terms of accuracy, which is the main practical shortcoming that hinders placing such solutions in practical use currently.
However, one must keep in mind that training FR on synthetic data is a very recently emerging research direction and it is already achieving higher recognition accuracies than solutions trained on synthetic data less than a decade ago \cite{DeepFace}. 

\subsection{Evaluating FR}
The need for large-scale FR evaluation datasets that represent real scenario variations is the main motivation for future research directions on synthetic data for FR evaluation. Although DGMs can generate arbitrary realistic face images, the utility of the generated images for FR remains challenging. Future research works include but are not limited to, DGMs for generating multiple faces of existing authentic identities, which might target specific variations such as age and pose, and generating complete evaluation datasets of multiple images of multiple identities.

\subsection{Attacking FR}
Even though creating novel attacks on identity management systems and society in general sounds is a serious malicious action, it is essential to foresee attacks created by real attackers to better enable their detection.
As the attackers would ask, the researchers should also ask ``What is the strongest attack I can create to serve the attack goals given the current state of basic technology?''
This follows the never-ending game of cat and mouse between attacks and attack mitigation.
Therefore, the constant struggle here is to always try to foresee new attacks and attack generation methodologies and analyze their strengths and weaknesses, leading to better mitigation strategies.

\subsection{Privacy enhancement}
The main challenge to generative face privacy enhancement is the generalizability and robustness as it must possess to maintain operation in real-world applications.
This generalization must ensure that the de-identification properties are strongly maintained even with unknown FR solutions.
The same goes for soft-biometric privacy, where the privacy-enhanced images should maintain their privacy properties when processed by diverse soft-biometric estimators with different levels of knowledge \cite{DBLP:conf/biosig/TerhorstHDRKSK20}.
Other open issues that still require increasing attention are the lack of clear quantifiability and provability privacy enhancement, the limited public benchmarks, and the need for controllable privacy where the user can have a choice of the privatised information \cite{DBLP:journals/tifs/MedenRTDKSRPS21}.

\subsection{Evaluation protocols}
We provided in this work an initial discussion on what synthetic data is needed for different FR use-cases and what properties are needed from such data based on the way it is used.
However, this initial discussion should evolve into a much-needed set of evaluation metrics and protocols that can precisely and comparably answer the question of ``How well does the created data fit its targeted properties within its use-case?''
Besides, and based on, the needed academic efforts in this regard, given that the synthetic data is foreseen to be a commodity, there is a need for such protocols and metric standards on the industrial level.
A clear candidate to develop such a standard would be the ISO SC37 work group 5 on Biometric testing and reporting.

\section{CONCLUSION}
The use of authentic data in FR poses technical, legal, and ethical concerns. 
However, such data plays a major role in training, evaluating, enhancing the FR user privacy, and even attacking FR.
This work provided initial discussions on the use of synthetic data in FR as an alternative to authentic data. We started by analysing and defining taxonomies for different possible FR use-cases in which synthetic data can be used. Then, we discussed the needed properties of synthetic data under each FR use-case. This has been followed by presenting the current state of synthetic FR. Finally, we provided several interesting directions of work that can be investigated in the future.
As a concluding remark, the use of synthetic data in different FR uses-cases is still in the early research stage and this work provides a base discussion on this research direction and aims at motivating and promoting further research works toward responsible FR development.     
 
\section*{Acknowledgment}
This research work has been funded by the German Federal Ministry of Education and Research and the Hessian Ministry of Higher Education, Research, Science and the Arts within their joint support of the National Research Center for Applied Cybersecurity ATHENE and the ARRS research program P2-0250 (B) Metrology and Biometrics Systems. This work has been partially funded by the German Federal Ministry of Education and Research (BMBF) through the Software Campus Project.

\bibliographystyle{elsarticle-num} 
\bibliography{main}

\end{document}